# A Dynamic-Neighbor Particle Swarm Optimizers for Accurate Latent Factor Analysis

Jia Chen
School of Cyber Science and Technology,
Beihang University
Beijing, China
chenjia@buaa.edu.cn

Yuanyi Liu
School of Cyber Science and Technology,
Beihang University
Beijing, China
YuanyiLiu@buaa.edu.cn

Xianchun Yi
School of Cyber Science and Technology,
Beihang University
Beijing, China
yixianchun@buaa.edu.cn

Renyu Zhang*
School of Cyber Science and Technology,
Beihang University
Beijing, China
18373072@buaa.edu.cn

Yang Hu*
School of Cyber Science and Technology,
Beihang University
Beijing, China
19373292@buaa.edu.cn

*Abstract*—High-Dimensional and Incomplete (HDI) matrices, which usually contain a large amount of valuable latent information, can be well represented by a Latent Factor Analysis (LFA) model. The performance of an LFA model heavily rely on its optimization process. Thereby, some prior studies employ the Particle Swarm Optimization (PSO) to enhance an LFA model's optimization process. However, the particles within the swarm follow the static evolution paths and only share the global best information, which limits the particles' searching area to cause sub-optimum issue. To address this issue, this paper proposes a Dynamic-neighbor-cooperated Hierarchical PSO-enhanced LFA (DHPL) model with two-fold main ideas. First is the neighbor-cooperated strategy, which enhances the randomly chosen neighbor's velocity for particles' evolution. Second is the dynamic hyper-parameter tunning. Extensive experiments on two benchmark datasets are conducted to evaluate the proposed DHPL model. The results substantiate that DHPL achieves a higher accuracy without hyper-parameters tunning than the existing PSO-incorporated LFA models in representing an HDI matrix.

## I. INTRODUCTION

In recent years, a large amount of High-Dimensional and Incomplete (HDI) matrices are generated and accumulated rapidly by the network services, industrial applications, and big-data-related information systems[1-2]. These HDI matrices contain many unknown useful latent information, and their data are extremely incomplete. This means the HDI matrices are difficult to extract and analyze, while they are worthy to analyze[3-7]. How to extract the latent information from HDI matrices and analyze them accurately has become a popular research issue[8-11].

According to previous studies[12-21], the Latent Factor Analysis (LFA) model process the HDI matrices efficiently. An LFA model first maps two high-dimensional entities involved in the target HDI matrix to a low-dimensional latent factor space. Second, it adopts the latent factors to construct a low rank approximation of the HDI matrix and corresponding objective function to measure the error between the original HDI matrix and its approximation. Third, the optimization algorithm is used to approximate the optimal latent factors by minimizing the objective function. Previous research works have demonstrated that the stochastic gradient descent (SGD) algorithm can optimize the latent factors in an LFA model accurately[22-24]. Some classical algorithms are proposed to improve the learning rate self-adaption of SGD algorithm[25-28]. Duchi et al.[25] propose the AdaGrad algorithm to adjust the learning rate by calculating the sum squares of gradients. Zeiler et al.[26] propose the AdaDelta algorithm, which uses the decaying average of all past squared gradients to adjust the learning rate. García-Galán et al.[28] propose the Adam algorithm to adjust the learning rate with the exponentially decaying average and the exponentially decaying square average of past stochastic gradients. These algorithms can effectively optimize the latent factors adaptively via improving the SGD algorithm. However, these optimization process costs much more time for adjusting the learning rate at each iteration.

To address this issue, Luo et al.[3] incorporates the classical Particle Swarm Optimization (PSO) algorithm into the optimization process of an LFA model. The PSO algorithm has been proven as an effective method in processing the large-scale data[29-31]. Luo et al. propose a position-transitional PSO-based LFA model, which uses the dynamic PSO algorithm to adjust the SGD's learning rate. To further improve accuracy of the PSO-based LFA model, Chen et al.[29] propose a hierarchical PSO-LFA (HPL) model, which refines the latent factors have been optimized by previous model with a newly proposed Mini-batch PSO (MPSO) algorithm and achieves higher accuracy. However, when the PSO algorithm refines the latent factors, each particle evolves following its particular path and only the global best information shares within the swarm. This work mechanism limits the particle swarm's searching area and causes the sub-optimum issue.

This research is supported by the CAAIHuawei MindSpore Open Fund under Grant CAAIXSJLJJ-2021-035A. (Corresponding author: Y. Hu, R. Zhang)

To expand the searching area and share more information among the particles, this work proposes a <u>D</u>ynamic-neighbor-cooperated <u>H</u>ierarchical <u>PS</u>O-enhanced <u>L</u>FA (DHPL) model, which improves the accuracy and self-adaption of an HPL model. The paper makes the following contributions:

a) A neighbor-cooperated MPSO algorithm. The neighbor particle's velocities are randomly added into the MPSO's evolution, which injects the disturbance into the original particle update process.

b) Adjust the hyper-parameters in the MPSO algorithm dynamically. Linearly decreasing inertia weight algorithm is adoptd to adjust hyper-parameters $\omega$, $\gamma_1$, and $\gamma_2$ in MPSO.

Empirical studies on two HDI matrices demonstrate that the proposed DHPL model outperforms the state-of-the-art LFA models.

This paper is structured as follows. Section II gives the preliminaries. Section III shows the methods. Section IV provides and analyses the empirical results. Finally, section V summarizes the algorithm proposed and discusses future work.

## II. Preliminaries

In this section, we first define the notations and formulas required in the paper. Then, we recall the fundamental models adopted in the paper.

### A. Notation

The notations used in this paper are listed in Table I below.

### B. Problem Setup

#### 1) An HDI Matrix

We use $U$ and $I$ to denote the large entity sets of users and items. Let us assume that $R^{|U|\times|I|}$ represents a HDI matrix when $|R_\Lambda| \ll |R_\Gamma|$. Given an entry $r_{u,i}$ denote the relationship value, where $u \in U$ and $i \in I$.

#### 2) An LFA Model

We assume that $\hat{R}=PQ^T$ denotes the estimation of low-dimensional latent factors for R. The latent feature matrices of $U$ and $I$ are defined by $P_{|U|\times f}$ and $Q_{|I|\times f}$, respectively. Euclidean distance is the objective function to estimate the difference between $\hat{R}$ and $R$. Furthermore, in order to improve its generalization ability and mitigate the magnitude effect, this work integrates a regularization term and linear bias. Therefore, an objective function is given as follow:

$$\sigma(P,Q,\mathbf{b},\mathbf{c}) = \frac{1}{2}\sum_{r_{u,i}\in R_\Lambda}(r_{u,i}-\sum_{k=1}^{f}p_{u,k}q_{k,i}-b_u-c_i)^2 \\ +\frac{\lambda}{2}\sum_{r_{u,i}\in R_\Lambda}(\sum_{k=1}^{f}p_{u,k}^2+\sum_{k=1}^{f}q_{k,i}^2+b_u^2+c_i^2), \quad (1)$$

where $b_u$ and $c_i$ represent linear biases for $u$ and $i$, $f$ represents the dimension of LF space, respectively [32, 33]. The regularization coefficient parameter is $\lambda$.

According to (1), the latent factors {$P$, $Q$, $\mathbf{b}$, $\mathbf{c}$} are updated iteratively with SGD[34, 35] as follows:

$$\forall r_{u,i} \in R_\Lambda, k \in (1,2,...,f): \\ \begin{cases} p_{u,k}^m = p_{u,k}^{m-1} - \eta \cdot \nabla \sigma_{u,i}(p_{u,k}^{m-1}), \\ q_{k,i}^m = q_{k,i}^{m-1} - \eta \cdot \nabla \sigma_{u,i}(q_{k,i}^{m-1}), \\ b_u^m = b_u^{m-1} - \eta \cdot \nabla \sigma_{u,i}(b_u^{m-1}), \\ c_i^m = c_i^{m-1} - \eta \cdot \nabla \sigma_{u,i}(c_i^{m-1}), \end{cases} \quad (2)$$

where $m$ and ($m$-1) denote the $m$-th and ($m$-1)-th iteration, $\eta$ represents the learning rate.

TABLE I. Notation Meaning and Description

| Notation | Description |
|---|---|
| $U$, $I$ | Involved entity sets. |
| $u$, $i$ | An element of $U$ and $I$. |
| $R$, $r_{u,i}$ | An $|U|\times|I|$ HDI matrix and its single element. |
| $\hat{R}$, $\hat{r}_{u,i}$ | $R$'s rank-$f$ approximation, and its single element. |
| $R_\Lambda$, $R_\Gamma$ | Known and unknown entry sets of $R$, and $|R_\Lambda|\ll|R_\Gamma|$. |
| $P$, $p_{u,f}$ | An $|U|\times f$ latent factor matrix for $U$, and its single element. |
| $\mathbf{p}_u$ | The $u$-th row vector of $P$. |
| $Q$, $q_{f,i}$ | A $f\times|I|$ latent factor matrix for $I$, and its single element. |
| $\mathbf{q}_i$ | The $i$-th row vector of $Q$. |
| $b, b_u$ | A vector of bias for $U$, and its $u$-th single element. |
| $c, c_i$ | A vector of bias for $I$, and its $i$-th single element. |
| $f$ | Dimension of the LF space. |
| $\sigma$ | A generalized loss function defined on $R_\Lambda$. |
| $\lambda$ | A regulation parameter for SGD-based LFA model. |
| $\eta$ | A learning rate parameter for SGD-based LFA model. |
| $V_1, V_2$ | Two particle vector sub-spaces. $V_1$ consists of latent factors in $P$ and $\mathbf{b}$, and $V_2$ consists of latent factors in $Q$ and $\mathbf{c}$. |
| $\mathbf{y}_k(n), \mathbf{l}_k(n)$ | The $k$-th particle's velocity, position at the $n$-th iteration. |
| $\mathbf{l}_{rd1}(n), \mathbf{l}_{rd2}(n)$ | The $k$-th particle's position of neighbors at the $n$-th iteration. |
| $\mathbf{y}_k^u(n), \mathbf{l}_k^u(n)$ | The $k$-th particle's velocity and position with the $u$-th parameter subset $[\mathbf{p}_u, b_u] \subset S_1$ at the $n$-th iteration. |
| $\mathbf{y}_k^i(n), \mathbf{l}_k^i(n)$ | The $k$-th particle's velocity and position with the $i$-th parameter subset $[\mathbf{q}_i, c_i] \subset S_2$ at the $t$-th iteration. |
| $\mathbf{p}_k^u(n), \mathbf{q}_k^i(n)$ | Position of $k$-th particle with $p_u$ in $V_1$, $q_i$ in $V_2$ at n-th iteration. |
| $\dot{\mathbf{p}}_k^u(n), \dot{\mathbf{q}}_k^i(n)$ | Velocity of $k$-th particle with $p_u$ in $V_1$, $q_i$ in $V_2$ at n-th iteration. |
| $b_k^u(n), c_k^i(n)$ | Position of $k$-th particle with $b_u$ in $V_1$, with $c_i$ in $V_2$ at the $t$-th iteration, respectively. |
| $\dot{b}_k^u(n), \dot{c}_k^i(n)$ | Velocity of $k$-th particle with $b_u$ in $V_1$, with $c_i$ in $V_2$ at the $t$-th iteration, respectively. |
| $K, k$ | Max number of particles and its $k$-th single element. |
| $D, d$ | Dimension number of particles, and its $d$-th single element. |
| $\gamma_1, \gamma_2, \gamma_3$ | Three acceleration coefficients for the PSO algorithm. |
| $r_1, r_2, r_3$ | Three uniform random parameters for PSO algorithm. |
| $\tilde{\mathbf{p}}_k$ | Historical best position vector of the $k$-th particle. |
| $\tilde{\mathbf{g}}$ | The best position vector in the particle swarm. |
| $\vec{P}, \vec{Q}$ | Vectorization of $P, Q$. |

## C. An HPL Model

The HPL model contains a two-layer structure and focuses on the second refinement layer. The second layer uses the pre-trained latent factors as input and refines them to achieve the higher accuracy, which solves the problem of premature convergence efficiently. The refinement algorithm is a Mini-Batch PSO (MPSO) algorithm which refines all the latent factors sequentially. For each latent factor, a specific particle swarm is constructed and optimized. The swarm is firstly constructed by mapping latent factors in two sub-vectors, which are defined as follows:

$$V = \left\{ \vec{P}, \mathbf{b}^T, \vec{Q}, \mathbf{c}^T \right\} \Rightarrow \begin{cases} V_1 = \left\{ \vec{P}, \mathbf{b}^T \right\}, \\ V_2 = \left\{ \vec{Q}, \mathbf{c}^T \right\}, \end{cases} \quad (3)$$

where $V$ is the particle vector space. It is divided into two mini-batch sub-vectors $V_1$ and $V_2$, respectively. With (3), the MPSO constructs the $|U|$ swarms containing $K$ particles to optimize each sub-vector $[\mathbf{p}_u, b_u], \forall u \in U$, and $|I|$ swarms with $K$ particles to optimize each sub-vector $[\mathbf{q}_i, c_i], \forall i \in I$, which are represented by the following equations:

$$\mathbf{l}_k^u(n) = \left[ \mathbf{p}_k^u, b_k^u \right], \mathbf{y}_k^u(n) = \left[ \dot{\mathbf{p}}_k^u, \dot{b}_k^u \right], \quad (4\text{-a})$$

$$\mathbf{l}_k^i(n) = \left[ \mathbf{q}_k^i, c_k^i \right], \mathbf{y}_k^i(n) = \left[ \dot{\mathbf{q}}_k^i, \dot{c}_k^i \right], \quad (4\text{-b})$$

where $\dot{\mathbf{p}}_k^u(n), \dot{b}_k^u(n)$ are the velocities of $k$-th particle of $\mathbf{p}_k^u(n)$ and $b_k^u(n)$ at the $t$-th iteration. $\dot{\mathbf{q}}_k^i(n)$ and $\dot{c}_k^i(n)$ are the velocities of $k$-th particle of $\mathbf{q}_k^i(n)$ and $c_k^i(n)$ at the $t$-th iteration. The specific $\mathbf{y}_k^u(n)$ and $\mathbf{l}_k^u(n)$ are updated with the PSO algorithm independently. The update equation can be denoted as:

$$\begin{cases} \mathbf{y}_k^u(n) = \omega \cdot \mathbf{y}_k^u(n-1) + \gamma_1 r_1 \left( \tilde{\mathbf{p}}_k(n-1) - \mathbf{y}_k^u(n-1) \right) \\ \qquad + \gamma_2 r_2 \left( \tilde{\mathbf{g}}(n-1) - \mathbf{l}_k^u(n-1) \right), \\ \mathbf{l}_k^u(n) = \mathbf{l}_k^u(n-1) + \mathbf{y}_k^u(n), \end{cases} \quad (5)$$

where $\tilde{\mathbf{p}}_k$ is historical best position vector of $k$-th particle, and $\tilde{\mathbf{g}}$ is the best position vector in the whole particle swarm. The following fitness function is used for $\mathbf{y}_k^u(n)$ and $\mathbf{l}_k^u(n)$:

$$F_u = \frac{1}{2} \sum_{r_{u,i} \in \Lambda(u)} (r_{u,i} - \mathbf{p}_u q_i - b_u - c_i)^2 + \frac{\lambda}{2} \left( \|\mathbf{p}_u\|^2 + |b_u|^2 \right). \quad (6)$$

With (6), the historical best position $\tilde{\mathbf{p}}_k^u(n)$ and the global best position $\tilde{\mathbf{g}}(n)$ at the $t$-th iteration is updated. With the above process, the swarm evolution for each $[\mathbf{p}_u, b_u]$ or $[\mathbf{q}_i, c_i]$ continues independently till $\tilde{\mathbf{g}}(n)$ converges or the maximum iteration count reaches. Based on the MPSO algorithm, the HPL model refines the latent factors with considerable time cost. In this paper, we design the dynamic-neighbor-cooperated algorithm and incorporate it into the HPL model, then use the linearly decreasing inertia weight strategy to adjust the hyper-parameters in the HPL model.

## III. PROPOSED APPROACH

We propose a DHPL model, which incorporates the dynamic neighbor-cooperated MPSO algorithm and linearly decreasing inertia weight strategy into HPL. In this section, we focus on the dynamic neighbor-cooperated MPSO algorithm and linearly decreasing inertia weight strategy.

### A. A Dynamic Neighbor-Cooperated MPSO Algorithm

In the standard PSO algorithm, each particle updates the current velocity and position based on its latest position and velocity, its historical best position, and the global best position in the particle swarm. A particle's current velocity and position don't only dependent on itself but is also influenced by the whole swarm. Inspired by the global information sharing strategy, we add the velocity information of neighbors into the velocity update formula of PSO algorithm. Use a specific row vector $u$ as an example, the update formulas for each particle's current velocity and position are presented as below:

$$\begin{cases} \mathbf{y}_k^u(n) = \omega \cdot \mathbf{y}_k^u(n-1) + \gamma_1 r_1 \left( \tilde{\mathbf{p}}_k(n-1) - \mathbf{l}_k^u(n-1) \right) \\ \qquad + \gamma_2 r_2 \left( \tilde{\mathbf{g}}(n-1) - \mathbf{l}_k^u(n-1) \right) \\ \qquad + \gamma_3 r_3 \left( \mathbf{l}_{rd1}(n-1) - \mathbf{l}_{rd2}(n-1) \right), \\ \mathbf{l}_k^u(n) = \mathbf{l}_k^u(n-1) + \mathbf{y}_k^u(n), \end{cases} \quad (7)$$

where $\mathbf{l}_{rd1}(n)$ and $\mathbf{l}_{rd2}(n)$ represent two randomly selected neighbors. We add their latest velocities into the velocity update formula. Thereby, the velocities of neighbors can be shared at each iteration and more global information are transfered to aovid the premature convergence at sub-optimum. Then, the particle's historical local best position is updated as follows:

$$\tilde{\mathbf{p}}_k(n) = \begin{cases} \mathbf{y}_k^u(n), \text{if } F\left(\mathbf{y}_k^u(n)\right) < F\left(\tilde{\mathbf{p}}_k(n-1)\right), \\ \tilde{\mathbf{p}}_k(n-1), \text{otherwise}. \end{cases} \quad (8)$$

The global best position can be updated after calculating all the particles' local best positions, which is formulated as:

$$\tilde{\mathbf{g}}(n) = \underset{\tilde{\mathbf{p}}_k(n)}{\arg\min} \left\{ F\left(\tilde{\mathbf{p}}_k(n)\right) \right\}. \quad (9)$$

Furthermore, each particle's velocity is limited in a proper range. Maximum and minimum values of the paticle's velocity are constrained as follows:

$$\mathbf{y}_k^u(n) = \begin{cases} \beta_{\max} \mathbf{l}_k^u(n), \text{ if } \mathbf{y}_k^u(n) > \beta_{\max} \mathbf{l}_k^u(n), \\ \beta_{\min} \mathbf{l}_k^u(n), \text{ if } \mathbf{y}_k^u(n) < \beta_{\min} \mathbf{l}_k^u(n), \end{cases} \quad (10)$$

where $\beta_{max}$ and $\beta_{min}$ denote the upper and lower ratio of velocity. In our work, the above dynamic neighbor-coorperated PSO updated formulas are incorporated into the MPSO algorithm, thereby constructing the DHPL model. The newly proposed DHPL model adopts the following two fitness functions:

$$F_1 = \sum_{r_{u,i} \in R_\Lambda(u)} (r_{u,i} - \mathbf{p}_u q_i - b_u - c_i)^2 + \lambda \left( \|\mathbf{p}_u\|^2 + |b_u|^2 \right), \quad (11\text{-a})$$

$$F_2 = \sum_{r_{u,i} \in R_\Lambda(u)} \left| r_{u,i} - p_u \mathbf{q}_i - b_u - c_i \right| + \lambda \left( \|\mathbf{q}_i\| + |c_i| \right), \quad (11\text{-b})$$

where $|\cdot|_{abs}$ represents the absolute value of a given value. We adopt the (11-a) as the fitness function for evaluation protocol RMSE, and (11-b) as the fitness function for evaluation protocol MAE.

### B. Linearly decreasing inertia weight strategy

The hyper-parameters in the MPSO algorithm are tunned manually, which can be updated with the proper algorithm. Because Shi et.al. has proven that the linearly decreasing weight strategy can promote the local earching ability[36]. We adopt the linearly decreasing weight strategy in our proposed DHPL model. We adjust the hyper-parameters with the linearly decreasing inertia weight strategy, which are formulated as:

$$\begin{cases} \omega = \omega_{\max} - (\omega_{\max} - \omega_{\min}) \cdot n / G, \\ \gamma_1 = \gamma_{\max} - (\gamma_{\max} - \gamma_{\min}) \cdot n / G, \\ \gamma_2 = \gamma_{\min} - (\gamma_{\max} - \gamma_{\min}) \cdot n / G, \end{cases} \quad (12)$$

where $\omega_{max}$ and $\omega_{min}$ are the maximal and minimal values of the inertia weight parameters, $G$ represents the maximum number of training iterations. $\gamma_{max}$ and $\gamma_{min}$ denote the maximal and minimal values of two acceleration coefficients.

## C. Algorithm Procedure and Time Cost analysis

Based on the above analysis, we design the DHPL model. As shown in Algorithm 1, we calculate the time cost of each row vector in the dynamic neighbor-coorperated MPSO algorithm, which can be calculated as follows:

$$C_1 = \Theta(|R_\Lambda| \times N \times K \times f). \tag{13}$$

Thereby, the time cost of the dynamic neighbor-coorperated MPSO algorithm can be calculated as the $M$ times of the optimization for each row vector and column vector, which can be calculated as follows:

| ALGORITHM: each row vector optimization | |
|---|---|
| **Input**: $U, I, R_\Lambda$ | |
| **Operation** | **Cost** |
| **Initialize:** $N$ = Maximum number of iterations | $\Theta(1)$ |
| **Initialize:** $P_{|U| \times f}, Q_{|I| \times f}, \mathbf{b}, \mathbf{c}$ | $\Theta(|U|+|I|+f)$ |
| **Initialize:** $\lambda, f, D = f + 1, K$ = Particle number | $\Theta(1)$ |
| **for** $u$ to $|U|$: | $\times |U|$ |
|    **Initialize:** $\tilde{\mathbf{p}}_k^u, \tilde{\mathbf{g}}^u$ | $\Theta(1)$ |
|    **Initialize:** $\mathbf{y}_k, \mathbf{l}_k$ | $\Theta(K \times D)$ |
|    **while** $n \leq N$ **and not** converge | $\times N$ |
|      **for** $k$=1 to $K$ | $\times K$ |
|        **for** $d$=1 to $D$ | $\times D$ |
|          Computing $\mathbf{y}_k, \mathbf{l}_k$ according to (7) | $\Theta(1)$ |
|        **end for** | -- |
|        **for** $u$=1 to $|R_\Lambda(U)|$ | $\times R_\Lambda(U)$ |
|          Computing $F_1(\mathbf{l}_{u\,k})$ according to (11) | $\Theta(f)$ |
|        **end for** | -- |
|        **if** $F_1(\mathbf{l}_k^u) < F_1(\tilde{\mathbf{p}}_k^u)$ | $\Theta(1)$ |
|          $\tilde{\mathbf{p}}_k^u = \mathbf{l}_k^u$ | $\Theta(1)$ |
|        **end if** | -- |
|        **if** $F_1(\mathbf{l}_k^u) < F_1(\tilde{\mathbf{g}}^u)$ | $\Theta(1)$ |
|          $\tilde{\mathbf{g}}^u = \mathbf{l}_k^u$ | $\Theta(1)$ |
|        **end if** | -- |
|      **end for** $k$ | -- |
|    **end while** | -- |
|    $[\mathbf{p}_u, \mathbf{b}_u] = \tilde{\mathbf{g}}^u$ | $\Theta(1)$ |
| **end for** | -- |
| **Output**: $P_{|U| \times f}, \mathbf{b}$ | |

$$C_{DN} = \Theta(|R_\Lambda| \times N \times K \times f \times M). \tag{14}$$

Based on (14), we have the time cost of the whole DHPL model as follows:

$$\begin{aligned} C_{DP-LFA} &= C_{PLFA} + C_{DN} \\ &= \Theta(|R_\Lambda| \times T_{PLFA} \times K_{PLFA} \times f) \\ &\quad + \Theta(|R_\Lambda| \times T \times K \times f \times M), \end{aligned} \tag{15}$$

where $T_{PLFA}$ and $K_{PLFA}$ represent the number of iterations and particles in PLFA model. From (15), it is obvious that the algorithm is linear in complexity and can be used in practice. Besides, its time cost is similar with the HPL model. The next section focuses on describing the experiments and analyzing the experimental results.

## IV. EXPERIMENTAL RESULTS AND ANALYSIS

### A. Experimental Settings

**Datasets.** ML10M[37] and Flixster[38] are chosen as the experimental datasets. We adopt 70%-10%-20% train-validation-test settings.

**Evaluation Protocol.** In this paper, we choose RMSE and MAE as our evaluation protocols. All experiments are performed on a MacBook pro with a 2.6 GHz Intel core CPU and 16 GB RAM. The formulas of RMSE and MAE are given as follows:

$$RMSE = \sqrt{\left(\sum_{r_{u,i} \in R_\Lambda}(r_{u,i} - \hat{r}_{u,i})^2\right) \Big/ |\tau|}, MAE = \left(\sum_{r_{u,i} \in R_\Lambda}|r_{u,i} - \hat{r}_{u,i}|\right) \Big/ |\tau|,$$

where $\tau$ represents the testing set disjoint with the training and validation sets.

**Model Settings.** We compared the DHPL model with various methods, including SGD-based LFA, Adam-based LFA,

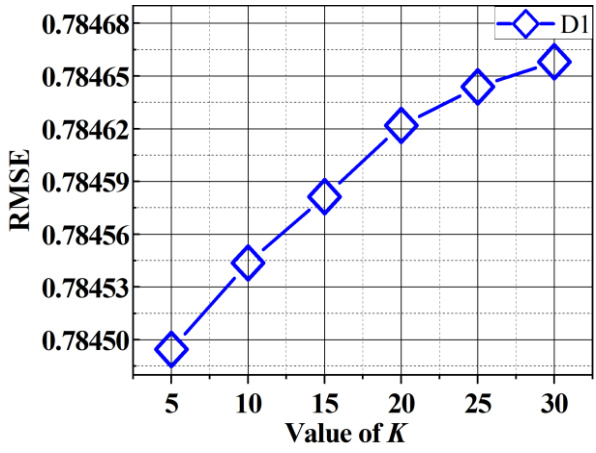 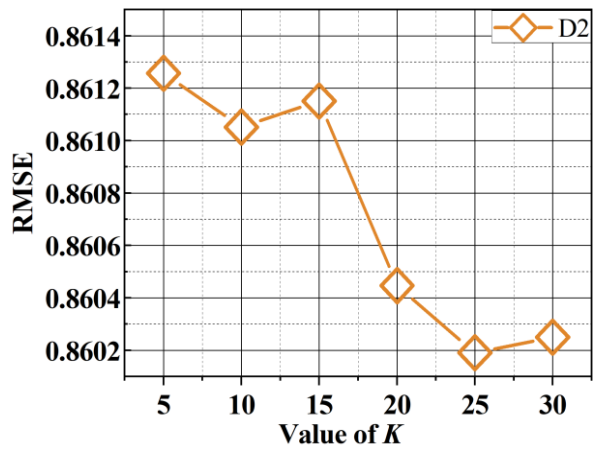

(a) RMSE on ML10M  (b) RMSE on Flixste

Fig. 1. RMSE of DHPL as *K* increases.

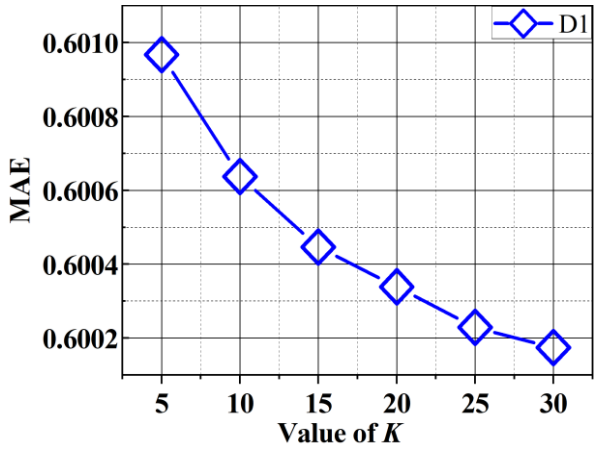 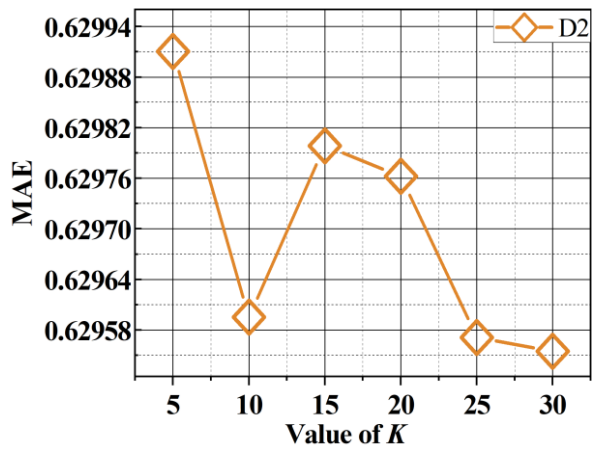

(a) MAE on ML10M  (b) MAE on Flixster

Fig. 2. MAE of DHPL as *K* increases.

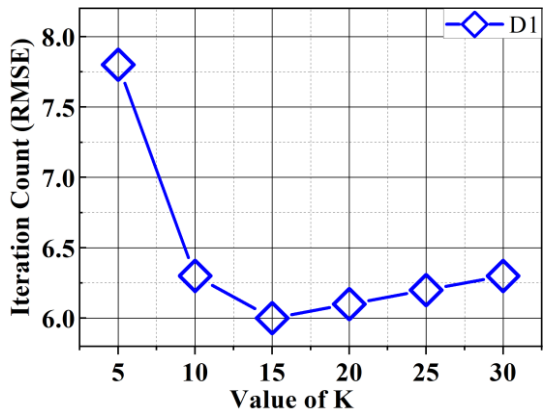 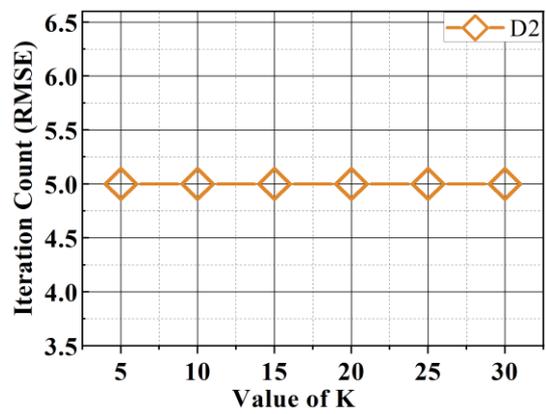

(a) Iterations on ML10M  (b) Iterations on Flixster

Fig.3. Converging iteration countS of DHPL as *K* increases.

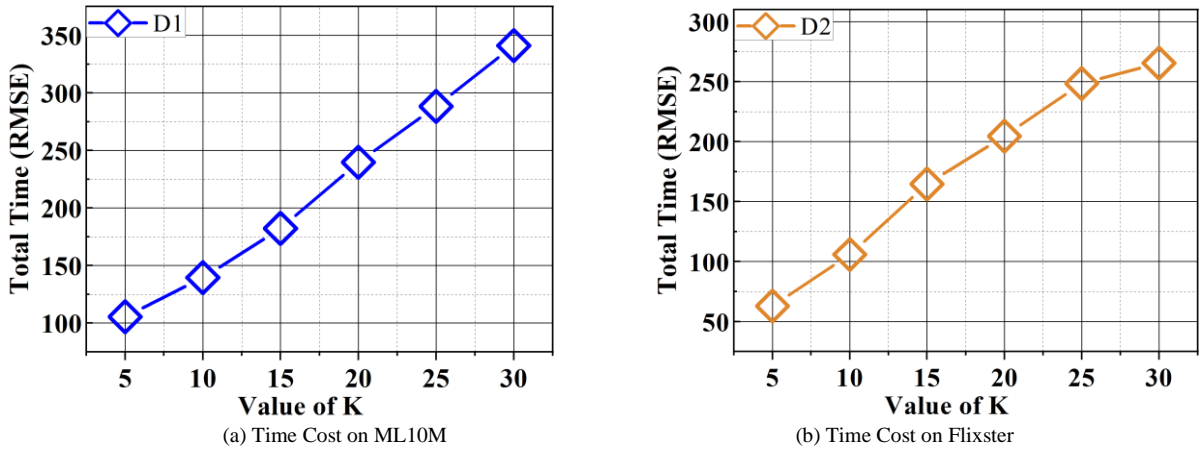

(a) Time Cost on ML10M  (b) Time Cost on Flixster

Fig. 4. Time cost of DHPL as *K* increases.

PLFA with biases and HPL. In this work, we set the hyper-parameters of Adam and HPL as the values has been widely used in previous studies. The regularization coefficient $\lambda$ is tuned for each dataset.

*B. Result comparison*

This sub-section compares the DHPL's performance with various swarm size, and the performance compared with other models.

*1) Comparison of swarm size*

The Swarm size *K* influences the performance of DHPL. We perform a series of experiments to analyze the performance of the DHPL model with $K \in \{5, 10, 15, 20, 25, 30\}$ on ML10M and Flixster. Fig. 1 and Fig. 2 depict DHPL's RMSE and MAE with various *K*, respectively. Fig. 3 and Fig. 4 depict the converging iteration counts with various *K*, respectively. For HPL and SGDE-PLFA, their involved PLFA model and the refinement process both converge with the same termination criterion as others, i.e., the error difference of two adjacent iterations is smaller than $10^{-4}$. From the results depicted from these figures, we conclude the following key findings:

a) **The accuracy of the DHPL model is insensitive to *K*.** In Fig. 1, the standard deviations of DHPL's RMSE on ML10M and Flixster are 6.30E-5 and 4.82E-4, respectively. In Fig. 2, the standard deviations of DHPL's MAE on ML10M and Flixster are 2.96E-4 and 1.45E-4, respectively.

b) **The DHPL model convergencs fast.** For instance, a DHPL model converges less than 8 iterations for RMSE with all the *K* values, which is depicted in Fig. 3(a). Obviously seen from Fig. 3(b), the DHPL model converges at 5 iterations with all the *K* values.

c) **Time cost varies linearly with the amount of *K*.** For instance, when RMSE is chosen, on ML10M, the time cost increases linearly from 105.39 to 341.13 seconds with *K* increasing from 5 to 30. On Flixster, the similar situation can be observed from Fig. 4(b).

TABLE II. COMPARISON RESULTS IN RMSE/MAE, INCLUDING WIN/LOSS COUNTS STATISTIC AND FRIEDMAN TEST

| Dataset | SGD-based LFA | Adam-based LFA | PLFA with Biases | HPL | DHPL |
|---|---|---|---|---|---|
| **ML10M** | 0.7872/ 0.6069 | 0.7901/ 0.6087 | 0.7854/ 0.6044 | 0.7851/ 0.5998 | **0.7845/ 0.6009** |
| **Flixster** | 0.9433/ 0.6653 | 0.9464/ 0.6661 | 0.8720/ 0.6512 | 0.8656/ 0.6394 | **0.8613/ 0.6299** |
| Win/Loss | 2/0 | 2/0 | 2/0 | 2/0 | -- |
| *F*-rank | 5 | 4 | 3 | 2 | 1 |

TABLE III. COMPARISON RESULTS IN TIME COST

| Dataset | SGD-based LFA | Adam-based LFA | PLFA with Biases | HPL | DHPL |
|---|---|---|---|---|---|
| **ML10M** | 874 | 891 | 157 | 249 | 262 |
| **Flixster** | 451 | 3393 | 91 | 222 | 154 |

Since the accuracy of DHPL is insensitive to *K*, and its CPU running time is linearly proportional to *K*, we select a small *K* value for the next experiments, i. e., five.

*2) Model Performance Analysis*

This sub-section compares the performance of DHPL with all the involved models. The lowest RMSE/MAE and Friedman Rank are recorded in Table II. Time cost is recorded in Table III. According to the experimental results, we have the following findings:

a) **A DHPL model's prediction is higher than that of its peers.** For instance, as shown in Table II, the RMSE with $K$=5 is 0.8613 on Flixster, which is 8.99% lower than Adam's 0.9464, 1.23% lower than PLFA's 0.8720, and 0.5% lower than HPL's 0.8656, respectively. Besides, the DHPL has the lowest $F$-rank value among its peers, which indicates it has the best prediction accuracy.

b) **A DHPL model's time cost is much lower than SGD-based model and Adam-based model, while its time cost is similar to HPL.** Obviously seen from Table III, the proposed DHPL consumes 262 seconds on ML10M, which is about 5.22% higher than HPL's 249 seconds. While on Flixster, DHPL consumes 154 seconds to converge, which is about 30.63% lower than HPL's 222 seconds. Note that the PLFA model is the first layer of both HPL and DHPL, thereby its time cost is the lowest among the involved models.

From the above analysis, we conclude that the DHPL model can achieve higher accuracy with a little more time.

## V. Conclusions

In this work, we propose a Dynamic-neighbor-cooperated PSO-enhanced LFA (DHPL) model. This model has two characteristics. One is incorporating the velocities of two randomly selected neighbors into the MPSO's iterative optimization process, the other is adopting the linearly decreasing inertia weight strategy to tune the hyper-parameters. With these two innovative approaches, the proposed DHPL model can expand the searching area and share more information among the particles, which increases the accuracy effectively. Empirical studies on two HDI matrices demonstrate that the proposed DHPL model outperforms the state-of-the-art LFA models.

In future, we will investigate the method of choosing the neighbors, discuss the influence on the number of neighbors, and incorporate some more state-of-the-art algorithms for tunning the hyper-parameters in DHPL model automatically.